\documentclass[runningheads,orivec]{llncs}

\usepackage{amsmath}
\usepackage{multirow}
\usepackage{threeparttable}
\usepackage[T1]{fontenc}
\usepackage{graphicx}
\usepackage{fontawesome}

\begin{document}

\title{TABLET: Table Structure Recognition using Encoder-only Transformers}

\author{Qiyu Hou\orcidID{0009-0009-4150-9907} \and Jun Wang\orcidID{0000-0002-9515-076X}(\faEnvelopeO)}
\authorrunning{Q. Hou et al.}
\institute{iWudao Tech\\
\email{\{houqy,jwang\}@iwudao.tech}}

\maketitle

\begin{abstract}
To address the challenges of table structure recognition, we propose a novel Split-Merge-based top-down model optimized for large, densely populated tables. Our approach formulates row and column splitting as sequence labeling tasks, utilizing dual Transformer encoders to capture feature interactions. The merging process is framed as a grid cell classification task, leveraging an additional Transformer encoder to ensure accurate and coherent merging.
By eliminating unstable bounding box predictions, our method reduces resolution loss and computational complexity, achieving high accuracy while maintaining fast processing speed.
Extensive experiments on FinTabNet and PubTabNet demonstrate the superiority of our model over existing approaches, particularly in real-world applications. Our method offers a robust, scalable, and efficient solution for large-scale table recognition, making it well-suited for industrial deployment.


\end{abstract}

\section{Introduction}

Tables, serving as a vital carrier of data, are prevalent across a wide range of digital documents. They efficiently store and display data in a compact and lucid format, encapsulating an immense volume of valuable information.
However, recognizing the structures of tables within digital documents, such as PDF files and images, and subsequently extracting structured data, present significant challenges due to the complexity and diversity of their structure and style~\cite{ICDAR-2024-iWudao}.

With the advancement of deep learning, new methodologies have surfaced, leading to significant progress in table structure recognition~\cite{ACMComputSurv2024_WHU,JIG2022_PKU}.
Recently, Large Vision-Language Models (LVLMs) with general capabilities have made significant progress in various document intelligence-related tasks, including table structure recognition~\cite{qwen25vl-Alibaba,InternVL25-Shanghai-AI-Lab}. We believe that, just as Large Language Models (LLMs) have gradually unified various NLP tasks, LVLMs will eventually unify a wide range of document intelligence tasks in the future.
However, at this stage, LVLMs still exhibit a noticeable accuracy gap compared to the current state-of-the-art methods in table structure recognition. Additionally, their high computational cost and relatively low processing speed present significant challenges for widespread industrial adoption.
This paper primarily focuses on the efficient processing of large-scale table data in real-world production environments, such as recognizing vast amounts of tabular data in various financial announcements. Therefore, we continue to concentrate on mainstream high-performance table structure recognition methods based on deep learning, which can be broadly categorized into three types: top-down, bottom-up, and end-to-end approaches.
Top-down approaches were among the first deep learning-based methods proposed for table structure recognition~\cite{DeepDeSRT_ICDAR-2017_DFKI}. These methods initially split the table into a fine-grained grid structure based on its row and column layout, then merge adjacent grid cells to reconstruct those spanning multiple columns or rows~\cite{SPLERGE_ICDAR-2019_Adobe}. While top-down approaches are generally reliable, both the splitting and merging processes involve complex post-processing, which can add computational overhead and potential challenges.
Meanwhile, a range of bottom-up approaches have also been introduced~\cite{TabStruct-Net_ECCV-2020_IIIT}. While these methods perform well on tables with clearly defined borders, they face significant challenges with borderless tables, where cell boundaries are less distinct. This issue is especially pronounced in sparse tables (i.e., those with many empty cells), as determining relationships between elements becomes more difficult due to the lack of visual and semantic cues in empty cells~\cite{LGPMA_ICDAR-2021_Hikvision}.
Recently, a series of end-to-end approaches based on image-to-sequence autoregressive models for generating structural sequences have gained increasing attention. These models directly output structural sequences along with cell or text coordinates~\cite{TableMASTER-HTML_arXiv2021_PingAn,TableFormer_CVPR-2022_IBM} or generate complete sequences in an OCR-free manner~\cite{OmniParser_CVPR-2024_Alibaba}. These approaches have demonstrated good performance, and due to their end-to-end nature, they eliminate the need for complex post-processing.
However, they also face inherent challenges due to the limitations of autoregressive language models. One major issue is that sequence-based autoregressive models cannot guarantee the validity and correctness of the output sequence (e.g., HTML). This is particularly problematic when encountering complex table headers that were not present in the training data, often leading to row-column misalignment. Furthermore, since errors typically occur early in the sequence—most often within the table headers—these errors propagate forward, causing increasing inaccuracies in the predicted coordinates of later table cells~\cite{VAST_CVPR-2023_Huawei}.  
Several studies have attempted to address these issues~\cite{VAST_CVPR-2023_Huawei,OTSL_ICDAR-2023_IBM,DRCC_IJCAI-2023_CAS,TFLOP_IJCAI-2024_Upstage-AI}, but they can only mitigate the problem rather than fully eliminate it. Additionally, for large and densely populated tables, the corresponding output sequences can be quite long, causing autoregressive methods to operate at a much slower speed.

Moreover, in real-world applications, the most commonly used metric, Tree-Edit-Distance-based Similarity (TEDS)~\cite{EDD_ECCV-2020_IBM}, inherently favors end-to-end autoregressive methods, which generate output cell by cell, while being relatively less favorable to top-down methods, which rely on splitting entire rows and columns. As a result, TEDS does not always fully reflect the true performance of top-down methods based on splitting and merging~\cite{TSRDet_arxiv2023_Ottawa}.
From our practical observations, top-down methods often prove to be more reliable and efficient, particularly in terms of achieving a higher proportion of fully correctly recognized tables. This makes them more suitable for deployment in industrial production environments. In response, a series of improved top-down approaches have emerged recently~\cite{SEMv2_PR2024_USTC,TSRFormer_ACM-MM-2022_Microsoft}. These methods enhance row and column feature representations by introducing a range of complex modules, thereby improving splitting and merging performance. Additionally, they have been designed to better adapt to tables in natural scenes or those appearing in curved and distorted images.

\begin{figure}
\includegraphics[width=\textwidth]{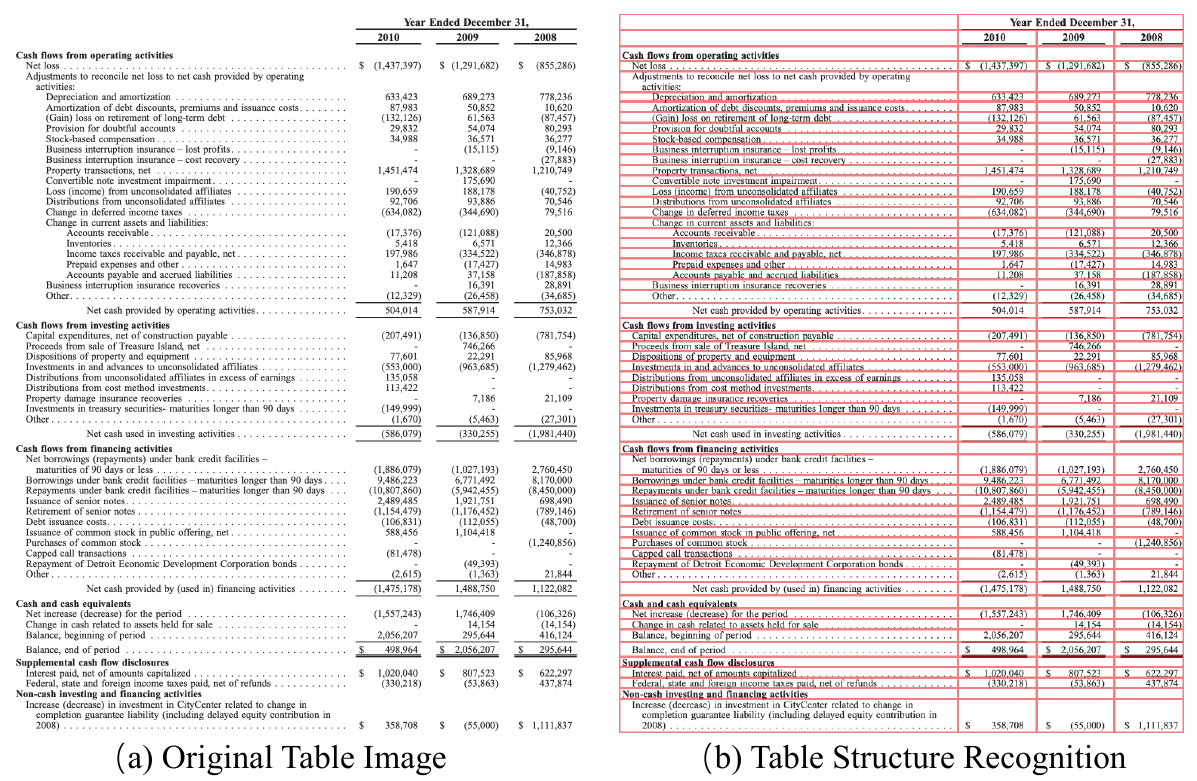}
\caption{An example of a densely populated table from a financial announcement and its corresponding table structure recognition.} \label{figure-dense-table-example}
\end{figure}

As previously mentioned, this paper primarily focuses on the efficient and reliable recognition of large volumes of tables in business documents, such as financial announcements. For large-scale document processing, a more effective strategy is to first apply document dewarping to eliminate distortions and noise before conducting layout analysis, rather than handling each type of page object separately. Consequently, our proposed method emphasizes structural complexity and does not rely on instance-level modules to recognize raw table images from diverse scenarios, in contrast to prior approaches~\cite{SEMv3_IJCAI-2024_USTC,TSRFormerDQ_PR2023_USTC}. Instead, we adopt ResNet~\cite{ResNet_CVPR-2016,ResNet_CVPR-2019} + FPN~\cite{CVPR-2017-FPN} to provide a more concise yet higher-resolution feature map, making it better suited for the large, densely populated tables with small fonts commonly found in financial documents as shown in Figure~\ref{figure-dense-table-example}.
After extracting horizontal and vertical feature sequences separately, our method avoids using complex one-dimensional DETR-based approaches for row and column segmentation. Instead, we employ a sequence labeling approach on horizontal and vertical feature sequences to split rows and columns. Unlike previous approaches that use RNN to model interactions within horizontal or vertical feature sequences~\cite{ICDAR-2021_PKU}, we leverage Transformer encoders, which capture interactions more efficiently and effectively.
During the merging process, the fine-grained grid cells obtained from the splitting step are processed using Region of Interest Align (RoIAlign)~\cite{ROI-Align_ICCV-2017} to extract grid cell-level features. A Transformer encoder is then applied to model interactions between grid cells, and a more concise categorization of the Optimised Table-Structure Language (OTSL)~\cite{OTSL_ICDAR-2023_IBM} is used to classify and merge grid cells, ultimately reconstructing the table structure.
We refer to the proposed method as \textbf{TABLET}: \textbf{TABLE} structure recognition using \textbf{E}ncoder-only \textbf{T}ransformers, which achieves both high performance and fast processing speed, making it well-suited for large-scale, real-world table recognition tasks.

This paper makes the following contributions:

$\bullet$ 
A new Split-Merge-based table structure recognition method TABLET is proposed, which formulates row and column splitting as feature sequence labeling tasks along their respective directions using dual Transformer encoders. The merging task is formulated as a grid cell classification problem using another Transformer encoder, where the cells obtained from the splitting process are classified to reconstruct the table structure.

$\bullet$ 
Our method TABLET features a succinct yet highly efficient model architecture, leveraging high-resolution feature maps to better handle the large, densely populated tables. Compared to autoregressive and bottom-up methods, our approach is more robust and achieves a higher percentage of fully and correctly recognized tables.

$\bullet$ 
Extensive experiments conducted on two widely used datasets, FinTabNet and PubTabNet, demonstrate the effectiveness and efficiency of our method compared to previous approaches.

\section{Related Work}

\subsection{Top-Down Approaches}

The top-down approaches follow a basic strategy of first detecting row and column separators to split the table into a grid structure, which is then merged to reconstruct cells spanning multiple columns or rows.
Early row-column splitting methods, such as DeepDeSRT~\cite{DeepDeSRT_ICDAR-2017_DFKI}, primarily relied on traditional object detection or semantic segmentation models to identify row and column regions separately. However, these methods failed to account for cells spanning multiple rows or columns, which limited their accuracy in handling complex table structures.
Some methods represent tables as one-dimensional feature sequences along the row and column directions, utilizing sequence labeling techniques to perform row and column splitting. SPLERGE~\cite{SPLERGE_ICDAR-2019_Adobe} is the first comprehensive approach. It first classifies pixels along the row and column directions to perform splitting and then determines relationships between adjacent grids to facilitate merging.
Some approaches incorporate RNN to better capture dependencies within the feature sequences~\cite{ICDAR-2019_NUST,ICDAR-2021_PKU}.
Subsequent research has introduced more complex methods to further enhance splitting and merging performance.
GrabTab~\cite{GrabTab_AAAI-2024_Tencent} introduces a deliberation mechanism and investigates its working patterns in component interaction for predicting complex table structures.
SEM~\cite{SEM_PR2022_USTC} preserves high-resolution pixel features, integrates semantic information, and utilizes GRU to learn relationships between grid units. SEMv2~\cite{SEMv2_PR2024_USTC} and SEMv3~\cite{SEMv3_IJCAI-2024_USTC} both employ Spatial CNN to capture spatial features. SEMv2 incorporates a Transformer to model and classify relationships between adjacent grids, while SEMv3 enhances features using row- and column-based self-attention, followed by OTSL-based classification and grid merging.
RobusTabNet~\cite{RobusTabNet_PR2023_USTC} was the first to employ Spatial CNN, laying the groundwork for subsequent advancements. TSRFormer~\cite{TSRFormer_ACM-MM-2022_Microsoft} built upon this by incorporating a DETR-based splitting method. Other methods have leveraged Transformer-based architectures to improve table structure recognition. TRUST~\cite{TRUST_arXiv2022_Baidu} employs a Transformer decoder for splitting, while Formerge~\cite{Formerge_ICDAR-2023_VTCC} utilize a Transformer encoder to learn and classify relationships between adjacent grids. DRCC~\cite{DRCC_IJCAI-2023_CAS} adopts a DETR-based approach, first splitting rows and then further generating cells autoregressively within each row, mitigating error accumulation. DTSM~\cite{DTSM_ICDAR-2024_SCUT} integrates text positional information, specifically designed to enhance the handling of densely populated tables.

\subsection{Bottom-Up Approaches}

A series of bottom-up approaches have also been proposed, which can be further categorized into cell-based methods and text block-based methods. 
Cell-based methods first detect the coordinates of individual cells and then determine their relationships, such as whether they belong to the same row or column, to reconstruct the table structure.
TabStruct-Net~\cite{TabStruct-Net_ECCV-2020_IIIT} employs object detection to locate individual cells within a table. It then extracts features from the detected cells and classifies cell pairs to determine whether they belong to the same row or column, ultimately reconstructing the complete table structure.
LGPMA~\cite{LGPMA_ICDAR-2021_Hikvision} focuses on cell detection, integrating local and global visual features to enhance table structure recognition. By leveraging pyramid mask re-scoring, it improves the reliability of cell alignment, ensuring greater detection accuracy. Additionally, it mitigates the impact of empty cells on detection models through soft pyramid mask learning.
LORE~\cite{LORE_AAAI-2023_ZJU} reconstruct table structures by regressing both spatial and logical positions, ensuring a more structured representation of the table layout.
GridFormer~\cite{GridFormer_ACM-MM-2023_Baidu} treats table cell intersections as vertices and cell boundaries as edges. It employs a DETR-like model to regress both vertices and edges, enabling a graph-based reconstruction of table structures.
In contrast, text block-based methods first detect the coordinates of text blocks and then infer their relationships (e.g., whether they belong to the same row, column, or cell) to reconstruct the table structure.
TIES~\cite{TIES-2.0_ICDAR-2019_NUST} was the first to employ a GNN to determine whether text blocks belong to the same row, column, or cell.
FLAG-Net~\cite{FLAG-Net_ACM-MM-2021_Tencent} introduced a flexible context aggregator, designed to classify text blocks based on row, column, and cell membership.
NCGM~\cite{NCGM_CVPR-2022_Tencent} presents a novel neural collaborative graph machine, which incorporates stacked collaborative blocks to alternately extract intra-modality context and model inter-modality interactions in a hierarchical manner.

\subsection{End-to-End Approaches}

The first notable work in this area was TableBank~\cite{TableBank_LREC-2020_Microsoft}, which employed an RNN-based autoregressive model to generate structural sequences for tables.
Similarly, TABLE2LATEX~\cite{TABLE2LATEX_ICDAR-2019_Harvard} generates LaTeX sequences, while EDD~\cite{EDD_ECCV-2020_IBM} produces HTML sequences.
In later developments, many methods adopted Transformer-based autoregressive models for table structure recognition. Notable examples include TableVLM~\cite{TableVLM_ACL-2023_Fudan}, MuTabNet~\cite{MuTabNet_ICDAR-2024_PFN}, SPRINT~\cite{SPRINT_ICDAR-2024_IIT-Bombay}, UniTabNet~\cite{UniTabNet_EMNLP-2024_USTC}, and approaches in~\cite{ICDAR-2023_NII,PRCV-2024_PKU}, among others.
In addition, here are some representative methods.
TableMaster~\cite{TableMASTER-HTML_arXiv2021_PingAn} employed Transformer decoders to generate HTML sequences, along with bounding boxes. TableFormer~\cite{TableFormer_CVPR-2022_IBM} introduced a lighter-weight network.
VAST~\cite{VAST_CVPR-2023_Huawei} proposed a visual alignment module to address inaccuracies in text bounding boxes.
TFLOP~\cite{TFLOP_IJCAI-2024_Upstage-AI} mitigates similar misalignment issues by incorporating OCR-extracted text content as input to the model.
Additionally, some methods leverage multi-task pre-trained models with instruction tuning for table structure recognition. Notable examples include OmniParser~\cite{OmniParser_CVPR-2024_Alibaba,OmniParserV2_arxiv2025_HUST}, TabPedia~\cite{TabPedia_NeurIPS-2024_USTC}, and PaliGemma2~\cite{PaliGemma-2_arXiv2024_Google}.

\section{System Framework}

\begin{figure}
\includegraphics[width=\textwidth]{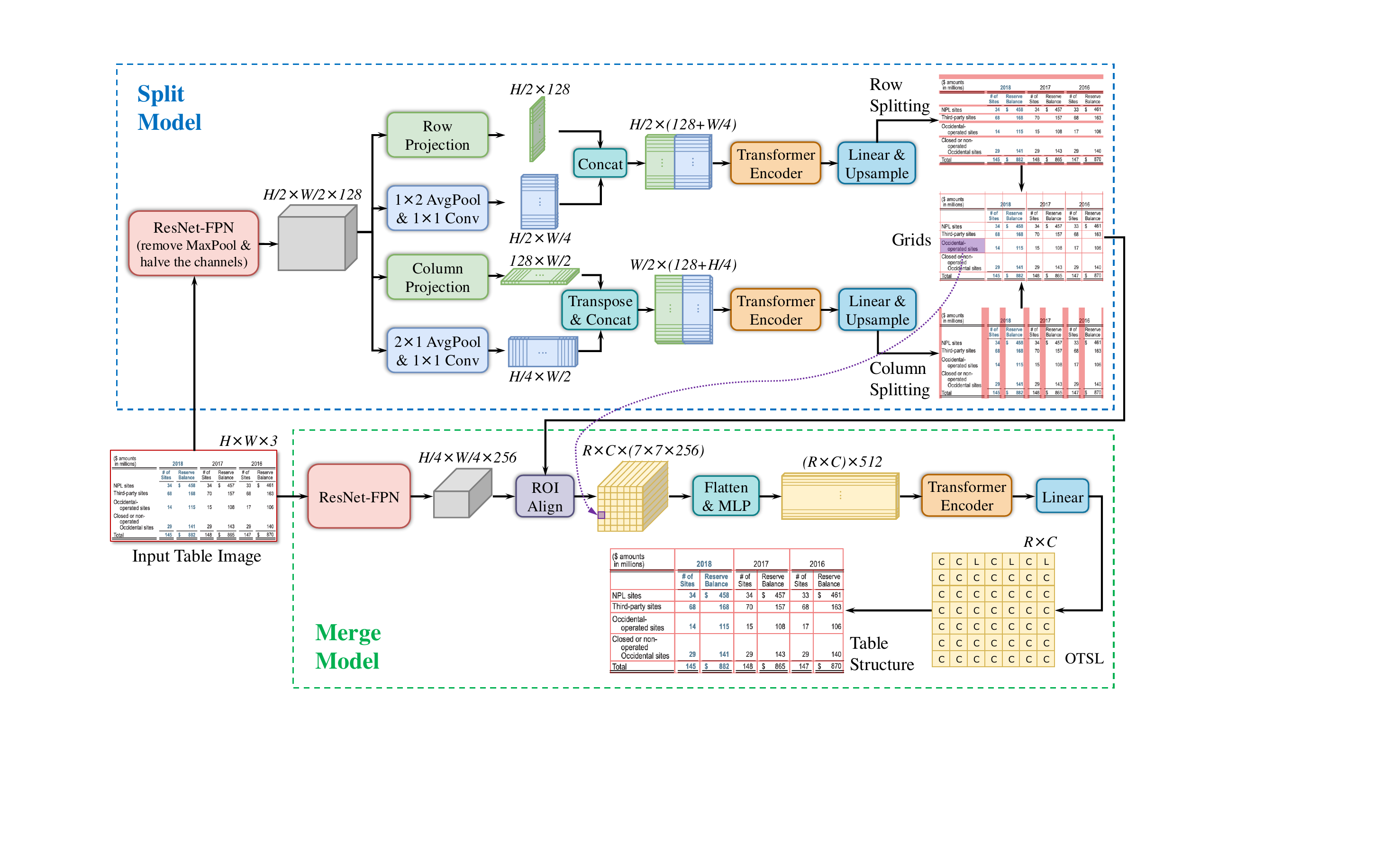}
\caption{System framework.} \label{System-Framework}
\end{figure}

As shown in Figure~\ref{System-Framework}, this section provides a brief introduction to the system framework of the proposed Split-Merge-based top-down method TABLET. In this framework, the split model and the merge model are trained separately.

Before being fed into the model, all table images undergo proportional scaling and blank padding to ensure a uniform size. Each image is then standardized to a fixed height of $H$ and a fixed width of $W$, resulting in a tensor of shape $ H \times W \times 3 $.

\subsection{Split Model}\label{split_model}

First, we use ResNet-18~\cite{ResNet_CVPR-2016,ResNet_CVPR-2019} combined with FPN~\cite{CVPR-2017-FPN} to extract multi-scale features. We observe that large, densely packed tables are prevalent in financial announcements and similar documents. Therefore, as in some previous works~\cite{SEM_PR2022_USTC}, we also consider it crucial to maintain the resolution of the feature map. To achieve this, we remove the Max Pooling layer from ResNet and halve the number of channels to reduce computational complexity, and combine it with a FPN having 128 channels. The resulting feature map, denoted as $F_{1/2}$, has a size of $H/2 \times W/2 \times 128$, making it half the resolution of the original input table image.

Inspired by SPLERGE~\cite{SPLERGE_ICDAR-2019_Adobe}, we further process $F_{1/2}$ to extract feature representations separately along the horizontal and vertical directions. This step is essential for the subsequent row splitting in the horizontal direction and column splitting in the vertical direction. The extracted features are composed of both global and local representations, ensuring a comprehensive understanding of the table structure.
In the horizontal direction, the global features are obtained by performing a projection along the horizontal axis, resulting in a feature map $F_{R_G}$ of size $H/2 \times 128$. Each element in $F_{R_G}$ is a learnable weighted average of the $W/2$ values in $F_{1/2}$ along the horizontal direction.  
The local features are extracted by first applying an Average Pooling operation with a kernel size of $1 \times 2$ to halve the width to $W/4$. Then, a $1 \times 1$ convolution is used to reduce dimensionality, producing a feature map $F_{R_L}$ of size $H/2 \times W/4$.
Next, $F_{R_G}$ and $F_{R_L}$ are concatenated, forming the final horizontal feature map $F_{R_{G+L}}$ of size $H/2 \times (128 + W/4)$.
In the vertical direction, the global features are obtained by projecting along the vertical axis, yielding a feature map $F_{C_G}$ of size $128 \times W/2$, where each element is a learnable weighted average of the $H/2$ values in $F_{1/2}$ along the vertical direction. 
The local features are extracted by first applying Average Pooling with a kernel size of $2 \times 1$, reducing the height to $H/4$, followed by a $1 \times 1$ convolution for dimensionality reduction. This results in a feature map $F_{C_L}$ of size $H/4 \times W/2$.
Finally, both $F_{C_G}$ and $F_{C_L}$ are transposed and concatenated to form the final vertical feature map $F_{C_{G+L}}$ of size $W/2 \times (128 + H/4)$.

Next, \(F_{R_{G+L}}\) and \(F_{C_{G+L}}\) are separately fed into two Transformer encoders~\cite{Transformer_NIPS-2017}.  
One Transformer encoder processes $F_{R_{G+L}}$ to capture relationships among feature vectors in the horizontal direction. This results in a feature map $F_R$ of the same size, enriched with context that enhances the final row splitting results.
The other Transformer encoder processes $F_{C_{G+L}}$ to capture dependencies among feature vectors in the vertical direction. This produces a feature map $F_C$, which provides richer context to improve the final column splitting results.  
Clearly, the input sequence lengths for the two Transformer encoders are fixed at $H/2$ and $W/2$, respectively. Each element in the sequence has a feature dimension of $128 + W/4$ for the horizontal encoder and $128 + H/4$ for the vertical encoder.  
Since these two encoders analyze 1-dimensional sequences along the horizontal and vertical axes, both use 1-dimensional positional embeddings, which are randomly initialized and learned during training.

The output at each position of the horizontal/vertical Transformer encoders is passed through a linear layer for binary classification, determining whether the line at the corresponding position in the horizontal/vertical direction belongs to a split region. 
Since the classification is performed on densely arranged lines in the horizontal and vertical directions, Focal Loss~\cite{Focal-Loss_ICCV-2017} is utilized for this task:
$FL_{split} = \frac{1}{n_h}  \sum_{i=1}^{n_h} \alpha_i (1 - p_i)^\gamma (-\log(p_i)) + \frac{1}{n_v}  \sum_{j=1}^{n_v} \alpha_j (1 - p_j)^\gamma (-\log(p_j))$.
\( n_h = H/2 \) represents the number of horizontal lines, and \( n_v = W/2 \) represents the number of vertical lines. Our experiments indicate that the best results are achieved by setting all class weights \( \alpha_i \) and \( \alpha_j \) to 1 and using a focusing parameter \( \gamma = 2 \) to smoothly control the downweighting of easy examples.  

Next, the classification results for each horizontal/vertical position are duplicated in place, effectively performing 2$\times$ upsampling. This results in $H$ classification outputs along the horizontal direction (matching the input image height) and $W$ classification outputs along the vertical direction (matching the input image width).
If a position (corresponding to a pixel) is classified as part of a split region, the corresponding horizontal or vertical line passing through that position can be used to draw the split region.
All text blocks within the table are first extracted using OCR. The center point of each text block is then projected along the horizontal/vertical directions. 
If any non-split regions in the horizontal/vertical direction contain no text projection, they are reclassified as split regions.

Similar to SPLERGE~\cite{SPLERGE_ICDAR-2019_Adobe}, after obtaining the final horizontal and vertical split regions, the midpoint of each split region is selected as the position for the splitting line.
The combination of horizontal and vertical split lines divides the table image into a grid structure with $R$ rows and $C$ columns.

\subsection{Merge Model}\label{merge_model}

Similarly, ResNet-18~\cite{ResNet_CVPR-2016,ResNet_CVPR-2019} combined with FPN~\cite{CVPR-2017-FPN} is used to extract multi-scale features. However, unlike the split model, a standard backbone is employed here.  
The resulting feature map, denoted as $F_{1/4}$, has a size of $H/4 \times W/4 \times 256$, making it one-fourth the size of the original input table image.

Inspired by SEMv3~\cite{SEMv3_IJCAI-2024_USTC} and Formerge~\cite{Formerge_ICDAR-2023_VTCC}, we use the grid structure obtained from the split model output to divide the feature map \( F_{1/4} \) accordingly.
For each grid cell, a $7 \times 7$ Region of Interest Align (RoIAlign)~\cite{ROI-Align_ICCV-2017} is applied to extract a $7 \times 7 \times 256$ feature map.  
The entire R-row, C-column grid forms a 3-dimensional feature map, denoted as $F_{grids}$, with a size of $R \times C \times (7 \times 7 \times 256)$.  

Next, $F_{grids}$ is flattened and passed through a two-layer MLP for dimensionality reduction, producing a 2-dimensional feature sequence $S_{grids}$ of size $(R \times C) \times 512$. Each grid cell in the table is represented by a 512-dimensional feature vector within this sequence, which has a total length of $R \times C$.

The feature vector sequence $S_{grids}$ is then fed into a Transformer encoder~\cite{Transformer_NIPS-2017} to capture relationships between grid cells.  
Unlike the split model, the merge model analyzes cells within a 2-dimensional grid. Therefore, similar to Formerge~\cite{Formerge_ICDAR-2023_VTCC}, it employs a learnable 2-dimensional positional embedding to encode spatial relationships effectively.

Next, a linear layer is applied to classify each grid cell based on the types defined in the Optimised Table-Structure Language (OTSL)~\cite{OTSL_ICDAR-2023_IBM}. This classification facilitates the merging of adjacent grid cells that need to be combined into table cells spanning multiple rows or columns.  
OTSL is designed to express table structures using a minimized vocabulary and a simple set of rules, significantly reducing complexity compared to HTML. Essentially, OTSL defines five tokens that directly represent a tabular structure based on an atomic 2-dimensional grid: ``C'' cell – A new table cell, which may or may not contain content. ``L'' cell – A left-looking cell that merges with its left neighbor to create a span. ``U'' cell – An up-looking cell that merges with its upper neighbor to create a span. ``X'' cell – A cross cell, merging with both the left and upper neighbors. ``NL'' (new-line) – Switches to the next row, commonly used in autoregressive methods for token sequence output.  
However, in Split-Merge methods, where the 2-dimensional grid is explicitly segmented, the ``NL'' category is unnecessary and thus not utilized.
The table images are split into densely arranged grid cells, so Focal Loss~\cite{Focal-Loss_ICCV-2017} is also used for the classification of these grid cells:
$FL_{merge} = \frac{1}{R \times C}  \sum_{k=1}^{R \times C} \alpha_k (1 - p_k)^\gamma (-\log(p_k))$.
\( R \times C \) represents the number of grid cells. Our experiments indicate that the best results are achieved by setting all class weights \( \alpha_k \) to 1 and using a focusing parameter \( \gamma = 2 \) to smoothly control the downweighting of easy examples.  
Finally, the OTSL representation is converted into HTML format. Then, based on their positions, the OCR-extracted text blocks are sequentially placed into their corresponding table cells.

\section{Experiments}

\subsection{Datasets and Metrics}

We conducted our experiments using two of the most widely used table structure recognition datasets. The first is FinTabNet~\cite{GTE-Cell_WACV-2021_IBM}: Version 1.0.0, which contains 91,596 tables in the training set, 10,656 in the validation set, and 10,635 in the test set. The second dataset is PubTabNet~\cite{EDD_ECCV-2020_IBM}: Version 2, with 500,777 tables in the training set, 9,115 in the validation set, and 9,138 in the test set.
These table datasets were typically annotated through automatic matching, which resulted in a considerable number of errors. Hou et al.~\cite{ICDAR-2024-iWudao} manually reviewed the FinTabNet test set, correcting and removing some erroneous table data, resulting in a refined FinTabNet test set consisting of 9,681 tables.
We also examined the test set of FinTabNet and found that some table structure annotations were incorrect or inconsistent. Corresponding examples can be found in the Appendix A~\ref{appendix_error}.
Table data is typically categorized into two types: simple and complex tables. The key distinction between them is whether they contain row-spanning or column-spanning cells.

All table images undergo preprocessing before being fed into the model proposed in this paper. Specifically, the longer side of each image is resized to 960 pixels while maintaining the original aspect ratio. Then, blank padding is added as needed to ensure that both the height ($H$) and width ($W$) are 960 pixels. For annotated split regions in the tables, if the original width is less than 5 pixels, the region is expanded based on its midpoint to ensure a minimum width of 5 pixels.

A variety of evaluation metrics have been used in previous table structure recognition studies. However, the most recent research predominantly adopts Tree-Edit-Distance-based Similarity (TEDS)~\cite{EDD_ECCV-2020_IBM} as the standard evaluation metric. The conventional TEDS metric assesses not only the accuracy of table structure recognition but also the quality of table cell content recognition. Additionally, a simplified version, TEDS-Struc~\cite{LGPMA_ICDAR-2021_Hikvision}, focuses solely on table structure recognition without considering the accuracy of table cell content.
Additionally, we consider an intuitive yet crucial metric: Accuracy, which measures the proportion of tables where both the structure and content are fully and correctly recognized. This metric is particularly important for critical domains that demand high precision.

\subsection{Implementation Details}

All experiments are implemented using PyTorch v2.2.2 and conducted on a machine equipped with two NVIDIA A100 80GB GPUs. The split model contains 16.1M parameters and the merge model 32.5M parameters.

In the split model, the backbone for basic image feature extraction is a modified ResNet-18 + FPN, where the number of channels in ResNet is reduced by half, and the number of channels in FPN is set to 128. Two Transformer encoders are used for row splitting and column splitting, both having the same architecture with 3 layers and 8 attention heads. As shown in Section~\ref{split_model}, the input sequence length is 480, with each element in the sequence having a dimensionality of 368. The feedforward network has a dimension of 2048, and the dropout rate is set to 0.1.

In the merge model, the backbone for image feature extraction is a standard ResNet-18 + FPN, where the number of channels in FPN is 256. The RoIAlign output size is 7 × 7. A two-layer MLP, composed of Linear + ReLU, is used, with both output and hidden layers set to 512-dimensional. The Transformer encoder has 3 layers and 8 attention heads. The input sequence length is 640, which is sufficient to cover a large number of grid cells in dense tables. As shown in Section~\ref{merge_model}, each element in the sequence has has 368 dimensions, while the feedforward network has a dimension of 2048, and the dropout rate is set to 0.1.

Both the split model and merge model are optimized using AdamW~\cite{AdamW_ICLR-2019} with a batch size of 32. The initial learning rate, betas, epsilon, and weight decay are set to 3e-4, (0.9, 0.999), 1e-8, and 5e-4, respectively.
The gradient norm is clipped using L2 normalization with $max\_norm = 0.5$. Additionally, the merge model adopts a polynomial decay schedule with a decay power of 0.9 for learning rate adjustment.
The split model and merge model are trained for 16 epochs and 24 epochs, respectively.

\subsection{Experimental Results and Analysis}

As previously discussed, this paper focuses on the accurate and high-speed processing of large-scale tables found in financial announcements and other business documents in real-world production environments. Therefore, we first conduct a detailed experimental analysis on the FinTabNet dataset.
FinTabNet provides PDF files of the pages containing tables, along with the corresponding bounding boxes for the tables, making it a dataset of overall higher quality compared to PubTabNet.
Table~\ref{table:FinTabNet_results} summarizes recent methods that use only images as model input and have been trained and evaluated on the FinTabNet dataset.
We observed that some of the experimental results reported in the table mistakenly mix up the validation and test sets of FinTabNet. For clarifications and additional details regarding this issue, please refer to the Appendix B~\ref{appendix_Clarification_FinTabNet}.

\begin{table}
\centering
\caption{The evaluation results on FinTabNet test set. Note: ``*'' indicates results reproduced or calculated by us.}\label{table:FinTabNet_results}
\begin{tabular}{l|ccc|ccc|c}
\hline
\multirow{2}{*}{Methods} & \multicolumn{3}{c|}{TEDS} & \multicolumn{3}{c|}{TEDS-Struc} & \multirow{2}{*}{Accuracy} \\
\cline{2-7}
& Simple & Complex & All & Simple & Complex & All & \\
\hline
TableMaster~\cite{TableMASTER-HTML_arXiv2021_PingAn} & - & - & 97.19* & - & - & 98.32* & 82.34* \\
TableFormer~\cite{TableFormer_CVPR-2022_IBM} & - & - & - & 97.50 & 96.00 & 96.80 & 77.98* \\
VAST~\cite{VAST_CVPR-2023_Huawei} & - & - & \underline{98.21} & - & - & 98.63 & - \\
TSRFormer(DQ-DETR)~\cite{TSRFormerDQ_PR2023_USTC} & - & - & - & - & - & 98.40 & - \\
Ly et al.~\cite{ICDAR-2023_NII} & - & - & 95.74 & - & - & 98.85 & - \\
GridFormer~\cite{GridFormer_ACM-MM-2023_Baidu} & - & - & - & - & - & 98.63 & - \\
UniTable~\cite{UniTable_NeurIPS-Workshop-2024_Georgia-Tech} & - & - & - & - & - & \underline{98.89} & - \\
MuTabNet~\cite{MuTabNet_ICDAR-2024_PFN} & - & - & 97.69 & - & - & 98.87 & - \\
SPRINT~\cite{SPRINT_ICDAR-2024_IIT-Bombay} & - & - & - & \underline{98.35} & \underline{97.74} & 98.03 & - \\
Zhu et al.~\cite{PRCV-2024_PKU} & - & - & 96.82 & - & - & \textbf{99.04} & - \\
BGTR~\cite{BGTR_ICPR-2024_SCUT} & - & - & - & - & - & \underline{98.89} & - \\
\hline
TABLET(Validation set) & 99.17 & 97.98 & 98.61 & 99.28 & 98.21 & 98.78 & 89.23 \\
TABLET(Test set) & \textbf{98.97} & \textbf{98.14} & \textbf{98.54} & \textbf{99.10} & \textbf{98.35} & 98.71 & \textbf{88.18} \\
TABLET(Refined set~\cite{ICDAR-2024-iWudao}) & 99.24 & 98.54 & 98.87 & 99.34 & 98.67 & 98.99 & 90.00 \\
\hline
\end{tabular}
\end{table}

Both image-to-sequence autoregressive and bottom-up table structure recognition methods involve predicting the bounding box locations of table cells. However, there is often a misalignment between the predicted cell positions and the corresponding text positions obtained through OCR.
In Table~\ref{table:FinTabNet_results}, some methods report only TEDS-Struc, which ignores content matching within table cells, rather than providing the full TEDS evaluation results.
For methods that report both TEDS and TEDS-Struc, a noticeable gap often exists between the two metrics for image-to-sequence autoregressive approaches. For instance, in the cases of Zhu et al.~\cite{PRCV-2024_PKU} and Ly et al.~\cite{ICDAR-2023_NII}, the gaps exceed 2 and 3, respectively, clearly reflecting the misalignment issue mentioned earlier. On the other hand, VAST~\cite{VAST_CVPR-2023_Huawei} introduces specific improvements to mitigate misalignment, thereby reducing this gap to some extent.
For methods based on the Split-Merge mechanism, misalignment between the positions of table cells and the positions of their contents rarely occurs. As a result, TEDS and TEDS-Struc scores are very close, as seen in our proposed methods, TABLET.
As previously discussed, the statistical methodology of TEDS and TEDS-Struc tends to favor autoregressive image-to-sequence methods. In contrast, Split-Merge-based methods are generally more robust and better suited for deployment in real-world production environments. 
In terms of Accuracy—the percentage of tables that are fully and correctly recognized—TableMaster~\cite{TableMASTER-HTML_arXiv2021_PingAn} and TableFormer~\cite{TableFormer_CVPR-2022_IBM} achieve only $82.34\%$ and $77.98\%$, respectively, whereas our proposed method exceeds $88\%$, demonstrating a significant improvement. Although the TEDS scores of TableMaster~\cite{TableMASTER-HTML_arXiv2021_PingAn} and TableFormer~\cite{TableFormer_CVPR-2022_IBM} show a much smaller gap compared to our TABLET method, the difference in Accuracy is substantial.
Additionally, our method has been evaluated on both the original FinTabNet validation and test sets~\cite{GTE-Cell_WACV-2021_IBM}, as well as the refined FinTabNet test set~\cite{ICDAR-2024-iWudao}. As expected, the results on the refined dataset show a slight improvement. And our method, TABLET, outperforms all previous approaches in terms of TEDS.

\begin{table}
\centering
\caption{Evaluation on different numbers of Transformer layers in the split model.}
\label{table:transformer_layer_split}
\begin{tabular}{cc|ccc|ccc|c}
\hline
\multicolumn{2}{c|}{Layers}  & \multicolumn{3}{c|}{TEDS} & \multicolumn{3}{c|}{TEDS-Struc} & \multirow{2}{*}{Accuracy} \\
\cline{1-8}
Row & Column & Simple & Complex & All & Simple & Complex & All & \\
\hline
0 & 0 & 98.36 & 97.81 & 98.08 & 98.64 & 98.09 & 98.35 & 85.95 \\
1 & 1 & 98.65 & 97.87 & 98.25 & 98.87 & 98.14 & 98.49 & 87.09 \\
3 & 3 & \textbf{98.97} & \textbf{98.14} & \textbf{98.54} & \textbf{99.10} & \textbf{98.35} & \textbf{98.71} & \textbf{88.18} \\
6 & 6 & 98.47 & 97.87 & 98.16 & 98.66 & 98.11 & 98.38 & 86.35 \\
\hline
\end{tabular}
\end{table}

\begin{table}
\centering
\caption{Evaluation on different numbers of Transformer layers in the merge model.}
\label{table:transformer_layer_merge}
\begin{tabular}{c|ccc|ccc|c}
\hline
\multirow{2}{*}{layers} & \multicolumn{3}{c|}{TEDS} & \multicolumn{3}{c|}{TEDS-Struc} & \multirow{2}{*}{Accuracy} \\
\cline{2-7}
& Simple & Complex & All & Simple & Complex & All & \\
\hline
0 & 98.88 & 97.47 & 98.15 & 99.01 & 97.70 & 98.33 & 81.40 \\
1 & 98.89 & 98.03 & 98.44 & 99.03 & 98.24 & 98.62 & 86.83 \\
3 & \textbf{98.97} & \textbf{98.14} & \textbf{98.54} & \textbf{99.10} & \textbf{98.35} & \textbf{98.71} & \textbf{88.18} \\
6 & 98.93 & 97.78 & 98.33 & 99.06 & 97.99 & 98.51 & 85.02 \\
\hline
\end{tabular}
\end{table}

In Table~\ref{table:transformer_layer_split}, we analyze the number of layers used in the Transformer encoders for row splitting and column splitting within the split model. Similarly, in Table~\ref{table:transformer_layer_merge}, we examine the number of layers in the Transformer encoder of the merge model. Experimental results indicate that all Transformer encoders achieve the best performance when using a three-layer architecture.

\begin{table}
\centering
\caption{Comparison of inference speed across different methods.}
\label{table:FPS_comparison}
\begin{tabular}{l|r|l|r|r}
\hline
Methods & Image Size & GPU & FPS & Adjusted \\
\hline
SEM~\cite{SEM_PR2022_USTC} & - & NVIDIA Tesla V100 32GB & 1.94 & - \\
RobusTabNet~\cite{RobusTabNet_PR2023_USTC} & 1024 & NVIDIA Tesla V100 & 5.19 & 7.33 \\
TSRFormer~\cite{TSRFormer_ACM-MM-2022_Microsoft} & 1024 & NVIDIA Tesla V100 & 5.17 & 7.31 \\
TRUST~\cite{TRUST_arXiv2022_Baidu} & 640 & NVIDIA Tesla A100 64GB & 10.00 & 4.44 \\
SEMv2~\cite{SEMv2_PR2024_USTC} & - & NVIDIA Tesla V100 32GB & 7.30 & - \\
VAST~\cite{VAST_CVPR-2023_Huawei} & 608 & NVIDIA Tesla V100 & 1.38 & 0.69 \\
TSRFormer(DQ-DETR)~\cite{TSRFormerDQ_PR2023_USTC} & 1024 & NVIDIA Tesla V100 & 4.17 & 5.89 \\
OmniParser~\cite{OmniParser_CVPR-2024_Alibaba} & 1024 & - & 1.30 & - \\
DTSM~\cite{DTSM_ICDAR-2024_SCUT} & 500 & NVIDIA TITAN Xp & 1.12 & 0.49 \\
OmniParser V2~\cite{OmniParserV2_arxiv2025_HUST} & 1024 & - & 1.70 & - \\
\hline
TABLET & 960 & NVIDIA A100 80GB & \textbf{18.01} & \textbf{18.01} \\
\hline
\end{tabular}
\end{table}

Table~\ref{table:FPS_comparison} lists all table structure recognition methods that report their processing speed in Frames Per Second (FPS). Some methods do not provide details on how their FPS was measured. However, our FPS evaluation of TABLET includes the entire pipeline for processing and recognizing each table image, encompassing image resizing, Split-Merge model inference, cell text content matching, and post-processing to generate HTML.
Our FPS experiments were conducted on FinTabNet using a single NVIDIA A100 80GB GPU. For current table recognition tasks, both the V100 and A100 GPUs provide sufficient memory. When using FP32 precision, the A100 (19.5 TFLOPS) offers approximately $24\%$ higher computational performance compared to the V100 (15.7 TFLOPS).
After accounting for differences in image size and GPU performance, we estimated the adjusted FPS of other methods under the same computational settings as ours, as shown in the last column of Table~\ref{table:FPS_comparison}.
Our method demonstrates a significant speed advantage over previous approaches. In terms of relative speed, it is approximately 2.5$\times$ faster than the previous best, RobusTabNet~\cite{RobusTabNet_PR2023_USTC}, and up to 30$\times$ faster than the slowest reported method, DTSM~\cite{DTSM_ICDAR-2024_SCUT}. Generally, end-to-end image-to-sequence methods, such as VAST~\cite{VAST_CVPR-2023_Huawei}, are inherently much slower than our Split-Merge-based approach due to their autoregressive nature.

We also evaluated the cutting-edge vision-language model Qwen2.5-VL-7B~\cite{qwen25vl-Alibaba}. On FinTabNet, its TEDS score was over 10\% lower than ours; after fine-tuning, the gap narrowed to approximately 2\%. However, its FPS was more than 100$\times$ slower than our method, rendering it unsuitable for large-scale financial document processing in our practical applications.

We also analyzed the errors reported in the recognition results on the FinTabNet test set and identified several common patterns:

$\bullet$ 
Failure to correctly split adjacent columns – In some cases, two adjacent columns are not properly separated due to overlapping vertical projections of their respective text regions. This misalignment prevents the split model from accurately distinguishing between them. Examples and further discussion can be found in Appendix~\ref{appendix-overlap-col}.

$\bullet$ 
Misalignment between a column header and its corresponding content – The header of a column is completely misaligned with the content below it. Additional examples and discussion can be found in Appendix~\ref{appendix-misaligned-col-header}.

$\bullet$ 
Incorrectly splitting multi-line text within a single cell into multiple rows – Text that should belong to a single cell is mistakenly divided into separate rows. It is worth noting that this type of annotation error is commonly observed in FinTabNet dataset. More details are provided in Appendix~\ref{appendix-wrong-split-rows}.

$\bullet$ 
Issues with empty cells – These cases typically do not affect subsequent table information extraction. It is important to note that FinTabNet dataset exhibits inconsistencies in its annotation of empty cells. Further discussion is available in Appendix~\ref{appendix-empty-cells}.

$\bullet$ 
Correct recognition by our model but incorrect annotations in FinTabNet – In some instances, our model produces correct recognition results, but the corresponding FinTabNet annotations are incorrect. Examples and discussion can be found in Appendix~\ref{appendix-wrong-annotation}.

\begin{table}
\centering
\begin{threeparttable} 
\caption{The evaluation results on PubTabNet validation set.}\label{table:PubTabNet_results}
\begin{tabular}{l|ccc|ccc}
\hline
\multirow{2}{*}{Methods} & \multicolumn{3}{c|}{TEDS} & \multicolumn{3}{c}{TEDS-Struc} \\
\cline{2-7}
 & Simple & Complex & All & Simple & Complex & All \\
\hline
TableMaster~\cite{TableMASTER-HTML_arXiv2021_PingAn} & - & - & 96.18\tnote{*} & - & - & - \\
LGPMA~\cite{LGPMA_ICDAR-2021_Hikvision} & - & - & 94.60 & - & - & 96.70 \\
TableFormer~\cite{TableFormer_CVPR-2022_IBM} & 95.40 & 90.10 & 93.60 & \underline{98.50} & \underline{95.00} & 96.75 \\
TRUST~\cite{TRUST_arXiv2022_Baidu} & - & - & 96.20 & - & - & 97.10 \\
SLANet~\cite{SLANet_arXiv2022} & - & - & 95.89 & - & - & 97.01 \\
TSRNet~\cite{TSRNet_PR2022_CAS} & - & - & - & - & - & 95.64 \\
VAST~\cite{VAST_CVPR-2023_Huawei} & - & - & 96.31 & - & - & 97.23 \\
TSRFormer(DQ-DETR)~\cite{TSRFormerDQ_PR2023_USTC} & - & - & - & - & - & 97.50 \\
Ly et al.~\cite{ICDAR-2023_NII} & \underline{98.07} & 95.42 & 96.77 & - & - & - \\
DRCC~\cite{DRCC_IJCAI-2023_CAS} & - & - & \textbf{97.80} & - & - & \textbf{98.90} \\
GridFormer~\cite{GridFormer_ACM-MM-2023_Baidu} & - & - & 95.84 & - & - & 97.00 \\
UniTable~\cite{UniTable_NeurIPS-Workshop-2024_Georgia-Tech} & - & - & 96.50 & - & - & 97.89 \\
MuTabNet~\cite{MuTabNet_ICDAR-2024_PFN} & \textbf{98.16} & \underline{95.53} & 96.87 & - & - & - \\
SEMv3~\cite{SEMv3_IJCAI-2024_USTC} & - & - & \underline{97.30} & - & - & 97.50 \\
Zhu et al.~\cite{PRCV-2024_PKU} & - & - & 96.77 & - & - & \underline{98.82} \\
BGTR~\cite{BGTR_ICPR-2024_SCUT} & - & - & 96.57 & - & - & 97.63 \\
\hline
TABLET & 97.72 & \textbf{95.83} & 96.79 & \textbf{98.60} & \textbf{96.70} & 97.67 \\
\hline
\end{tabular}
\begin{tablenotes}
      \item[*] The authors of TableMaster updated the results in their paper and published them at https://github.com/JiaquanYe/TableMASTER-mmocr.
    \end{tablenotes}
\end{threeparttable} 
\end{table}

Finally, we also trained and evaluated our method on PubTabNet, another widely used dataset.
Unlike FinTabNet, PubTabNet does not provide the original PDF files from which the tables are extracted. Instead, it only offers relatively low-resolution table images and does not include precise text line information. As a result, obtaining the table's textual content requires using an OCR tool. Due to this limitation, the TEDS evaluations in Table~\ref{table:PubTabNet_results} may vary across different methods, as each method could use a different OCR tool, leading to inconsistencies in reported results.
For this study, we extracted table text using the OCR model specifically trained for PubTabNet by TableMaster~\cite{TableMASTER-HTML_arXiv2021_PingAn}. As shown in Table~\ref{table:PubTabNet_results}, our proposed method, TABLET, also achieves outstanding results on the PubTabNet dataset, performing on par with the current state-of-the-art methods.
DRCC~\cite{DRCC_IJCAI-2023_CAS} employs DETR for row coordinate regression. Although DETR is relatively challenging to train, the semi-autoregressive approach is less affected by resolution compared to the pixel classification-based Split-Merge method, resulting in strong evaluation performance.
Compared to other recent methods, our proposed method, TABLET, achieves comparable TEDS scores. However, when compared with TSRFormer~\cite{TSRFormerDQ_PR2023_USTC} and SEMv3~\cite{SEMv3_IJCAI-2024_USTC}, both of which also adopt the pixel classification-based Split-Merge mechanism, TABLET slightly outperforms them in terms of TEDS-Struc.
Notably, SEMv3~\cite{SEMv3_IJCAI-2024_USTC} achieves a high TEDS score of 97.3, even though TEDS-Struc is only 97.5, demonstrating that they used a more accurate OCR tool.

\section{Conclusion}
In this paper, we propose a succinct yet powerful table structure recognition method, TABLET, that reformulates row and column splitting as feature sequence labeling tasks along their respective axes. By harnessing dual Transformer encoders for horizontal and vertical sequence modeling and introducing an additional Transformer-based grid classification mechanism for merging, TABLET captures intricate structural dependencies with high accuracy and efficiency. The high-resolution feature maps further enable TABLET to handle complex, densely populated tables commonly found in real-world production environments. Experimental results on FinTabNet and PubTabNet confirm the efficiency and effectiveness of our approach, outperforming existing autoregressive and bottom-up methods.

\bibliographystyle{splncs04}
\bibliography{ref}

\section{Appendix A: Analysis of errors in the table structure recognition results}\label{appendix_error}

\subsection{Overlap of text regions between adjacent columns}\label{appendix-overlap-col}

\begin{figure}
\includegraphics[width=\textwidth]{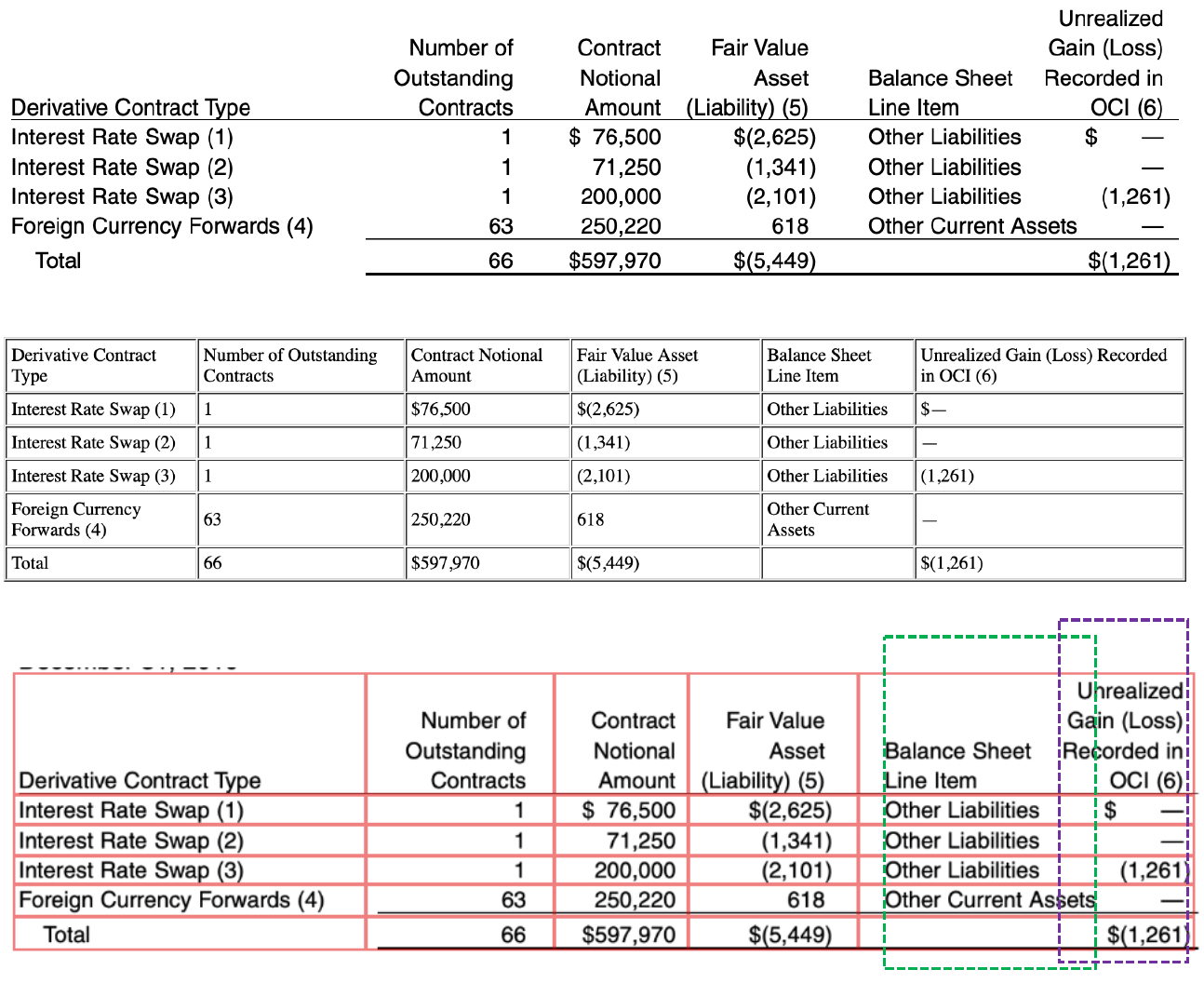}
\caption{An example of a recognition error caused by the overlap of text regions between adjacent columns.} \label{figure-split-error-example2}
\end{figure}

In Figure~\ref{figure-split-error-example2}, the correct table structure is presented using an HTML table. It can be observed that the second-to-last column and the last column were not correctly split. This issue occurs because the vertical projection area of the text in the second-to-last column (marked with a green bounding box) overlaps with the vertical projection area of the text in the last column (marked with a purple bounding box). As a result, the split model fails to correctly separate these two columns.

\subsection{A column header completely misaligned with its corresponding content below it}\label{appendix-misaligned-col-header}

\begin{figure}
\includegraphics[width=\textwidth]{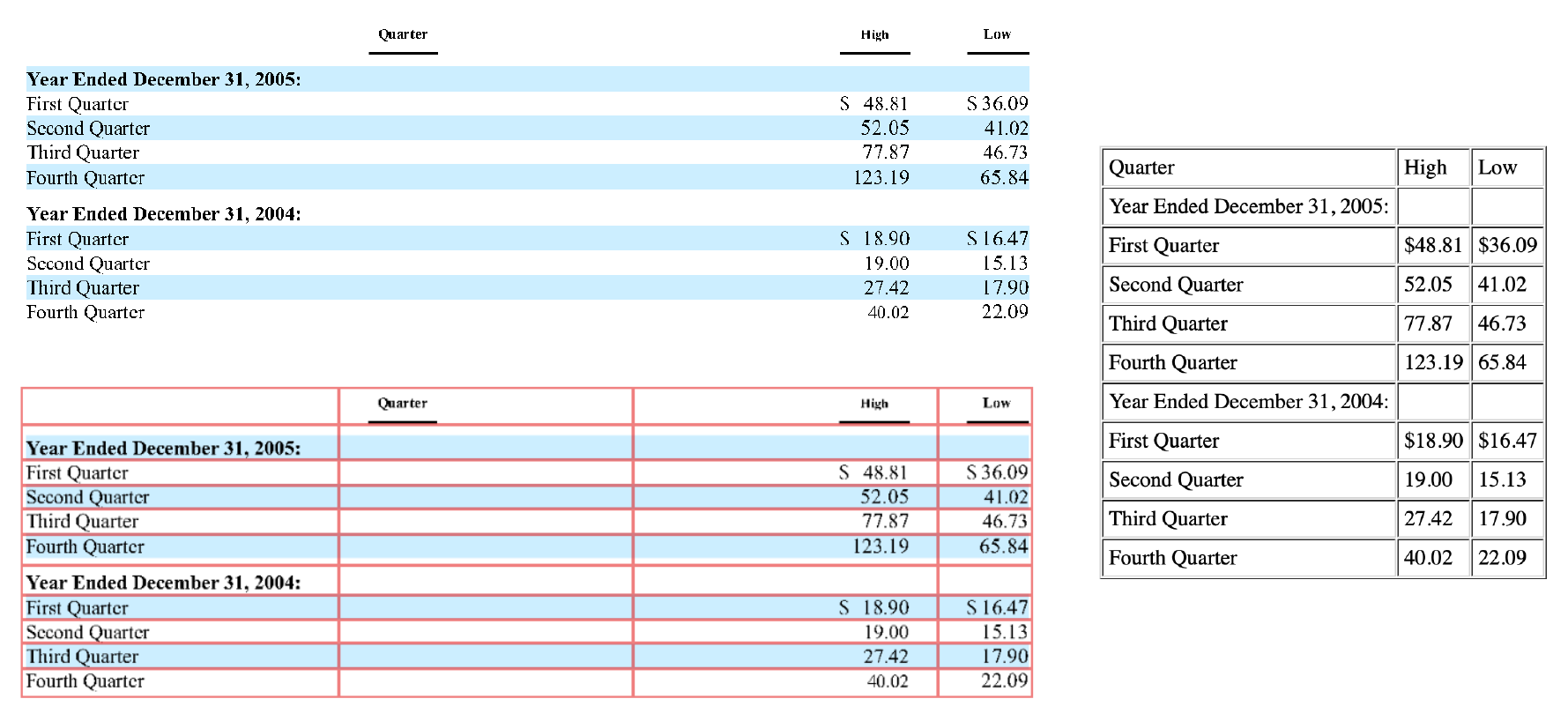}
\caption{An example of a column header completely misaligned with its corresponding content below it.} \label{figure-error-col-misalign}
\end{figure}

In Figure~\ref{figure-error-col-misalign}, the correct table structure is presented using an HTML table. It can be observed that the column header ``Quarter'' in the first column is completely misaligned with the rest of the content in that column. As a result, the split model mistakenly separates ``Quarter'' into a standalone column.

\subsection{Text that spans multiple lines within a single cell is incorrectly split into separate rows}\label{appendix-wrong-split-rows}

\begin{figure}
\includegraphics[width=\textwidth]{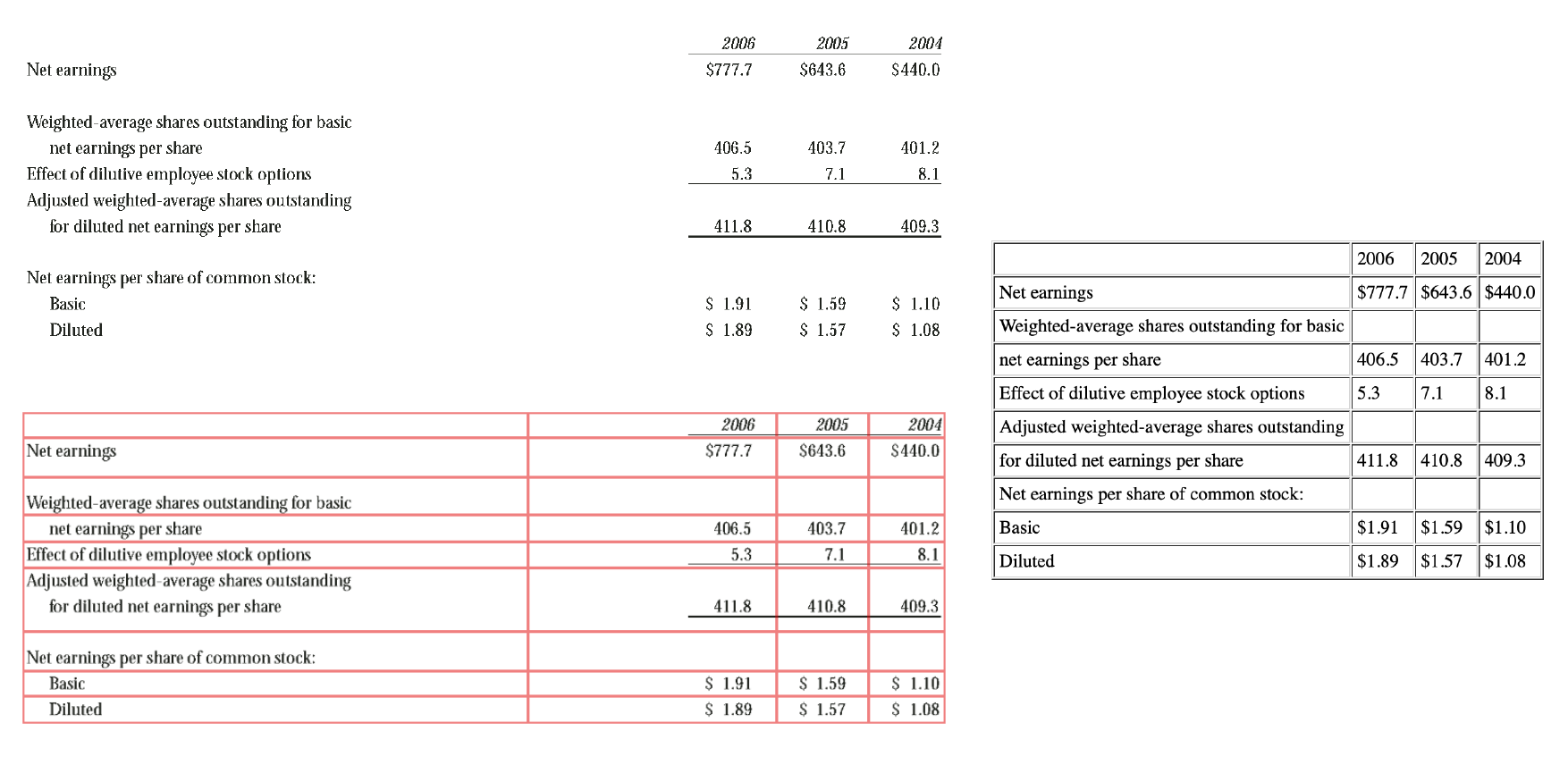}
\caption{An example where text spanning multiple lines within a single cell is incorrectly split into separate rows.} \label{figure-error-row-split}
\end{figure}

In Figure~\ref{figure-error-row-split}, the text ``Weighted-average shares outstanding for basic net earnings per share'' appears on two lines but should belong to a single cell rather than being incorrectly split into two rows. In such cases, relying solely on visual information can sometimes make accurate recognition challenging. Interestingly, the annotations provided by FinTabNet, when displayed in an HTML table, mistakenly separate multi-line text that belongs to a single cell into different rows as well.

\subsection{Issues with the annotation of empty cells}\label{appendix-empty-cells}

\begin{figure}
\includegraphics[width=\textwidth]{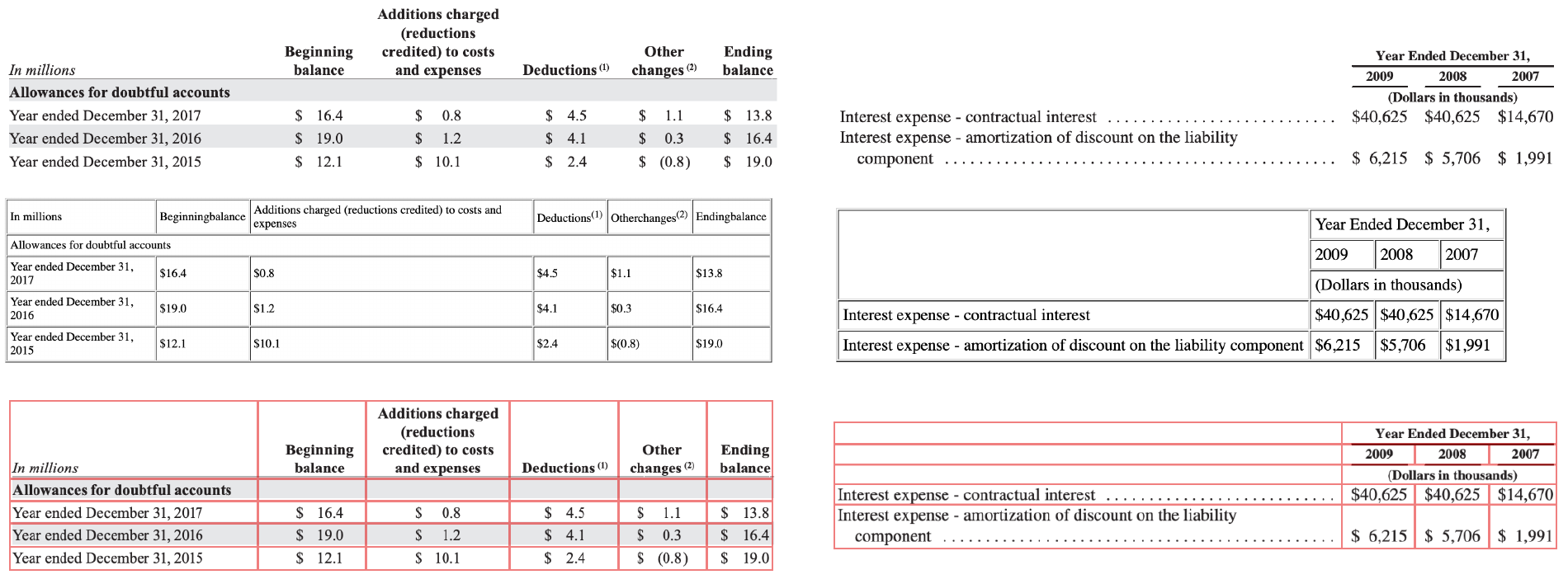}
\caption{Two examples of tables with empty cells.} \label{figure-issues-empty-cells}
\end{figure}

In Figure~\ref{figure-issues-empty-cells}, the two tables illustrate a discrepancy in the handling of adjacent empty cells. In the recognition results produced by our model, adjacent empty cells are not merged, whereas in the annotations provided by FinTabNet, displayed through HTML tables, these empty cells are merged.
It is important to note that whether these empty cells are merged or not does not impact the subsequent extraction of information from the table. Therefore, strictly speaking, the output of our model should not be considered incorrect. Additionally, FinTabNet exhibits inconsistencies in its annotation of empty cells. In most cases we observed, adjacent empty cells in the annotation data were not merged, making the merged representation, as shown in the HTML tables, an exception rather than the norm.

\subsection{Correct recognition results but incorrect annotations}\label{appendix-wrong-annotation}

As we previously mentioned, FinTabNet contains a significant number of annotation errors. When comparing the recognition results of our model with the annotated data in the FinTabNet test set, we found that many reported errors were actually due to incorrect annotations in the test set, while our model's recognition results were, in fact, correct.

\begin{figure}
\includegraphics[width=\textwidth]{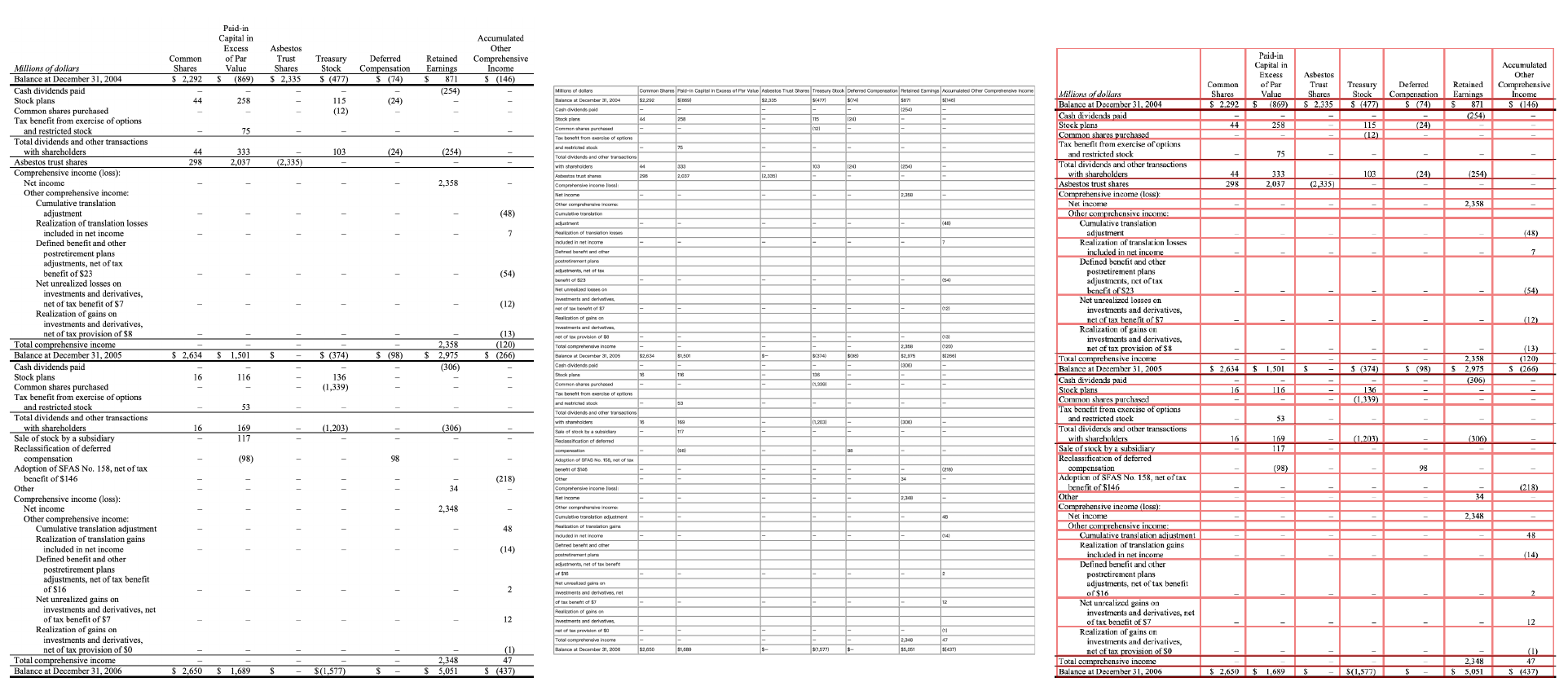}
\caption{Example 1 of correct recognition but incorrect annotations.} \label{figure-correct-wrong-split-1}
\end{figure}

\begin{figure}
\includegraphics[width=\textwidth]{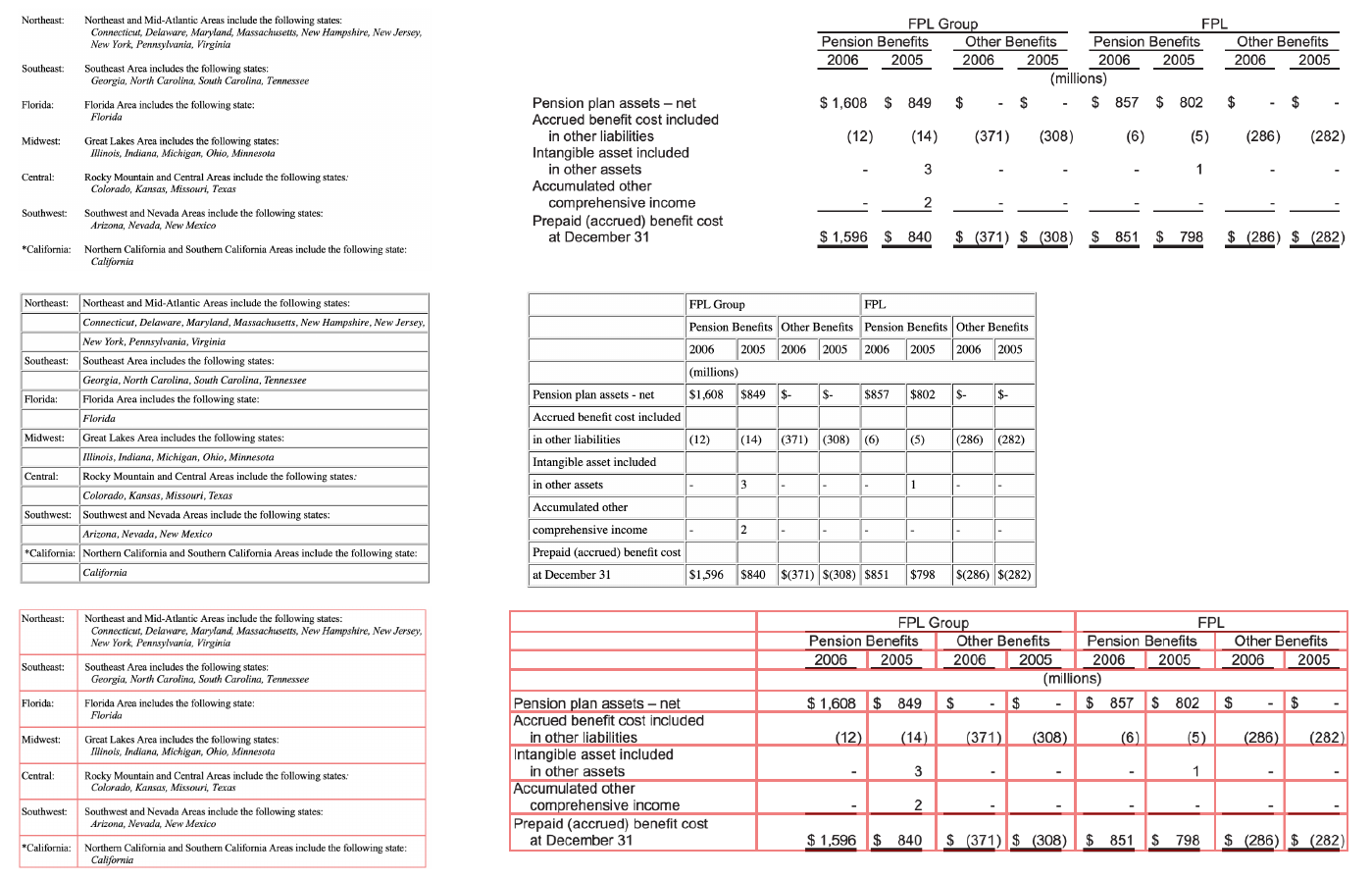}
\caption{Example 2 of correct recognition but incorrect annotations.} \label{figure-correct-wrong-split-2}
\end{figure}

In Figure~\ref{figure-correct-wrong-split-1} and Figure~\ref{figure-correct-wrong-split-2}, our model's recognition results are correct. However, the HTML tables provided by FinTabNet exhibit a common annotation error, where text that should belong to a single cell is incorrectly split into multiple rows.

\begin{figure}
\includegraphics[width=\textwidth]{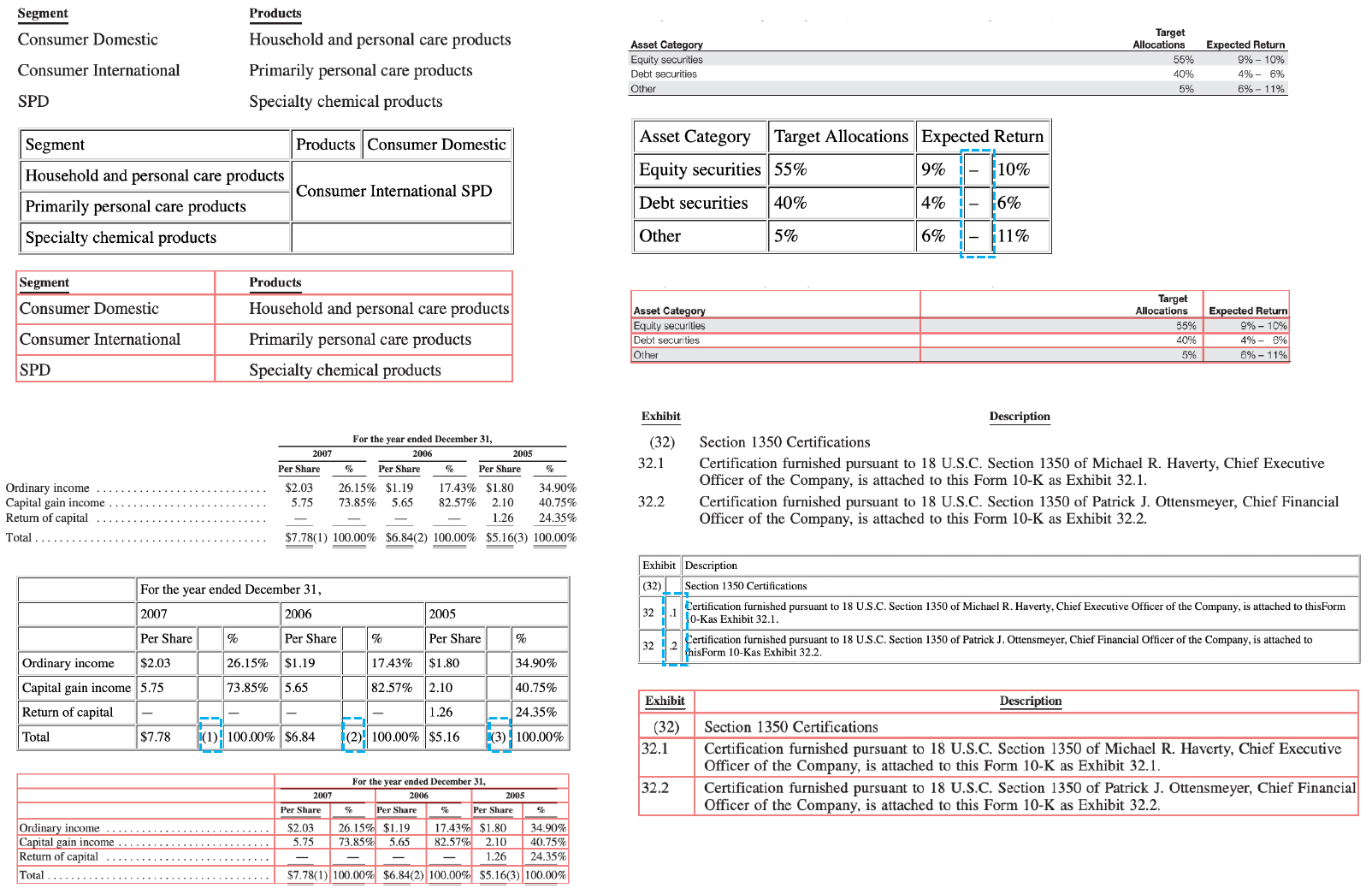}
\caption{Example 3 of correct recognition but incorrect annotations.} \label{figure-correct-wrong-mis-1}
\end{figure}

\begin{figure}
\includegraphics[width=\textwidth]{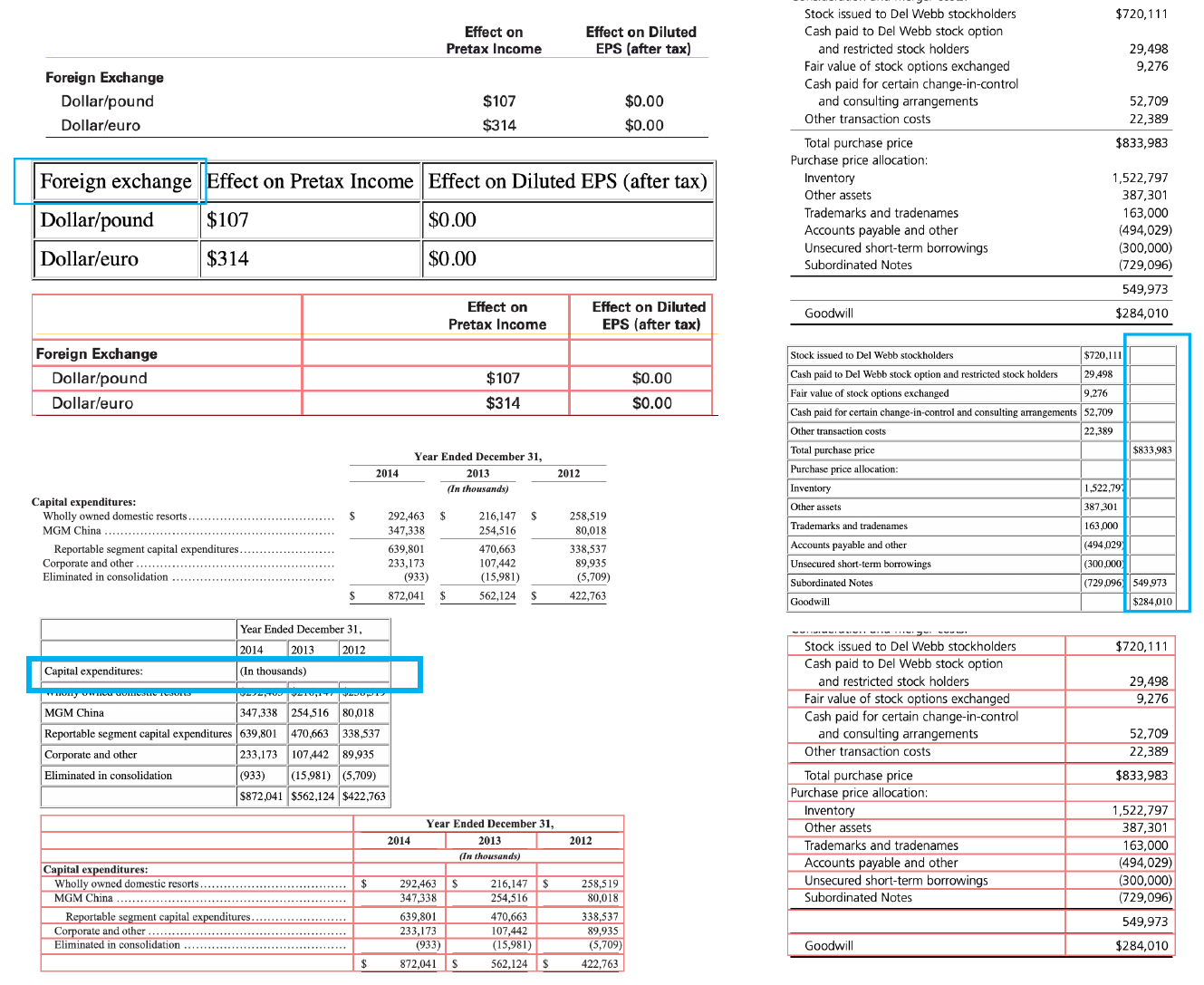}
\caption{Example 4 of correct recognition but incorrect annotations.} \label{figure-correct-wrong-mis-2}
\end{figure}

In Figure~\ref{figure-correct-wrong-mis-1} and Figure~\ref{figure-correct-wrong-mis-2}, our model's recognition results are correct. However, the HTML tables presented here highlight various misalignment annotation errors in FinTabNet.

\section{Appendix B: Clarification on the validation and test sets of FinTabNet}\label{appendix_Clarification_FinTabNet}

In the paper introducing and describing FinTabNet~\cite{GTE-Cell_WACV-2021_IBM}, the reported number of tables in the validation and test sets are 10,635 and 10,656, respectively. However, after conducting a detailed comparison and analysis of the downloaded FinTabNet dataset, we found that the actual sizes should be the opposite: the validation set contains 10,656 tables, while the test set contains 10,635 tables. This discrepancy is likely a typo in the original paper.

The test set consists of 5,126 simple tables and 5,509 complex tables, while the validation set contains 5,698 simple tables and 4,958 complex tables. Notably, the validation set has a higher proportion of simple tables, which tend to yield higher evaluation scores.

For example, based on the data in Table~\ref{table:FinTabNet_results}, we can verify that SPRINT~\cite{SPRINT_ICDAR-2024_IIT-Bombay} was evaluated on the test set, as its reported score satisfies the following equation:
\[
98.35 \times 5126 + 97.74 \times 5509 = 98.03 \times 10635
\]

On the other hand, for TableFormer~\cite{TableFormer_CVPR-2022_IBM}, while Figure 5 and Figure 7 of the paper indicate that the test set was used, the evaluation results provided in Table 2 actually correspond to the validation set, as the following equation holds:
\[
97.5 \times 5698 + 96.0 \times 4958 = 96.8 \times 10656
\]

\end{document}